%% file: ltexpprt_doublecolumn.tex
\documentclass[twoside,leqno,twocolumn]{article}

\usepackage[letterpaper]{geometry}
\usepackage{amsmath} 
\usepackage{amsfonts} 
\usepackage{ltexpprt}
\usepackage{hyperref}
\usepackage{cite}
\usepackage{amssymb}
\usepackage{graphicx}
\usepackage{textcomp}
\usepackage{comment}
\usepackage{hyperref}
\usepackage{multirow}
\usepackage{tabularray}
\usepackage{subfig}
\usepackage{algorithm}
\usepackage{algorithmicx}
\usepackage{algpseudocode}
\usepackage{cuted}
\usepackage{amssymb} 
\usepackage{etoolbox}
\usepackage{array} 
\usepackage[table,xcdraw]{xcolor}
\usepackage{tabularx}
\usepackage{booktabs}
\makeatletter
\patchcmd{\@begintheorem}{\textit}{\textbf}{}{}
\patchcmd{\@begindefinition}{\textit}{\textbf}{}{}
\makeatother

\begin{document}

\newcommand\relatedversion{}

\newcommand{\showcomment}{} 
\newcommand{\addcomment}[2]{\ifdefmacro{\showcomment}{{\textcolor{#1}{#2}}}{}}
\newcommand{\dl}[1]{{\addcomment{blue}{#1}}}
\newcommand{\dk}[1]{{\addcomment{red}{[DK: #1]}}}

\title{\Large Unveiling the Impact of Local Homophily on GNN Fairness: \\ In-Depth Analysis and New Benchmarks}
\author{Donald Loveland\thanks{University of Michigan, Ann Arbor \{dlovelan, dkoutra\}@umich.edu}
\and Danai Koutra\footnotemark[1]}

\date{}

\maketitle


\fancyfoot[R]{\scriptsize{Copyright \textcopyright\ 20XX by SIAM\\
Unauthorized reproduction of this article is prohibited}}





\begin{abstract} \small\baselineskip=9pt
Graph Neural Networks (GNNs) often struggle to generalize when graphs exhibit both homophily (same-class connections) and heterophily (different-class connections). Specifically, GNNs tend to underperform for nodes with local homophily levels that differ significantly from the global homophily level. This issue poses a risk in user-centric applications where underrepresented homophily levels are present. Concurrently, fairness within GNNs has received substantial attention due to the potential amplification of biases via message passing. However, the connection between local homophily and fairness in GNNs remains underexplored. In this work, we move beyond global homophily and explore how local homophily levels can lead to unfair predictions. We begin by formalizing the challenge of fair predictions for underrepresented homophily levels as an out-of-distribution (OOD) problem. We then conduct a theoretical analysis that demonstrates how local homophily levels can alter predictions for differing sensitive attributes. We additionally introduce three new GNN fairness benchmarks, as well as a novel semi-synthetic graph generator, to empirically study the OOD problem. Across extensive analysis we find that two factors can promote unfairness: (a) OOD distance, and (b) heterophilous nodes situated in homophilous graphs. In cases where these two conditions are met, fairness drops by up to 24\% on real world datasets, and 30\% in semi-synthetic datasets. Together, our theoretical insights, empirical analysis, and algorithmic contributions unveil a previously overlooked source of unfairness rooted in the graph's homophily information. Code available at \href{https://anonymous.4open.science/r/HeteroFairICDM-F7DE/README.md}{Repo}.\end{abstract}

\input{PAGES/introduction}
\input{PAGES/preliminaries}

\input{PAGES/theory}
\input{PAGES/datasets_and_synth_method}

\input{PAGES/results}
\input{PAGES/related_work}
\input{PAGES/conclusion}
\bibliographystyle{IEEEtran}
\bibliography{reference}
\input{PAGES/appendix}

\end{document}

%% file: PAGES/introduction.tex
\section{Introduction}
\label{intro}
Graph Neural Networks (GNNs) have achieved success in graph machine learning by enabling representation learning over graph-structured data \cite{wu2020survey}. 
However, it has been shown that GNNs do not always outperform non-graph baselines \cite{zhu2020h2gcn, yan2022twocoin}. To understand the conditions under which GNNs succeed, recent studies have investigated trends between graph structure and performance \cite{liu2023generalized, topping2022understanding, ma2022is}. In particular, the presence of homophily and heterophily -- where nodes tend to connect to similar and different classes, respectively -- has proved challenging to simultaneously model. Thus, extensive work has explored the conditions under which GNNs perform well for both homophilous and heterophilous graphs, often focusing on degree and distinguishability \cite{ma2022is}.

Originally used to describe the overall graph connectivity patterns, homophily has been expanded to characterize the local neighborhoods around nodes \cite{loveland2022graph, mao2023demyst}. This local perspective has helped refine the notion that GNNs perform poorly under heterophily, instead highlighting that a major difficulty lies in predicting for underrepresented local homophily levels, regardless of the global level \cite{loveland2023perf, mao2023demyst, du2022gbk}. 
We refer to the challenge of applying trained GNNs to settings with underrepresented local homophily levels as the out-of-distribution (OOD) local homophily problem. While existing research has primarily explored this issue from a performance perspective, we propose that the correlation, either causal or spurious, between sensitive attributes and class labels can allow the OOD problem to potentially compromise fairness in GNNs  \cite{mengan2020Fair, dulhanty2019auditing}. 
From a fairness standpoint, most studies have investigated how GNNs might exploit sensitive user attributes \cite{agarwal2021towards, wang2022fair, chen2024fairness}, with \textit{group fairness} measuring treatment differences between sensitive groups \cite{dong2023fairness}. However, these studies often assess disparity on a global scale, overlooking disparities induced by local graph structures  \cite{dai2021fairgnn, li2020dyadic, dong2022edits, rahman2019fairwalk, spinelli2021dropout}. Recognizing that local patterns in graphs may be influenced by harmful processes, such as forced segregation in social networks \cite{mcpherson2001birds}, we emphasize the need to formally link local homophily and fairness to ensure fair predictions. Without local characterization, GNNs risk misusing underrepresented sensitive attributes, constituting a novel source of unfairness rooted in the local structures of the graph.

\textbf{This Work.} Moving beyond a global interpretation of group fairness and homophily, we investigate \textit{local homophily-induced unfairness}, arising from a GNN's ability to exploit a node's sensitive attribute. Focusing on node classification, we motivate the problem through theoretical analysis and show that predictions for nodes with differing sensitive attributes can diverge in OOD settings. To generalize and validate our theoretical findings, we conduct empirical analyses using an OOD training framework, controlling the discrepancy in homophily level between the training and test sets. For datasets, we identify inadequacies in current benchmarks for studying the OOD problem due to limited homophily variability, insufficient preprocessing information, and weakly defined sensitive attributes. Consequently, we introduce three new GNN fairness benchmarks with diverse local homophily levels, easy reproducibility, and meaningful fairness tasks. To further explore various local homophily patterns, we propose a semi-synthetic graph generation strategy based on graph re-wiring for precise control.
Our theoretical insights, novel datasets, and algorithmic contributions provide evidence that  GNNs induce systematic group unfairness for individuals whose local homophily differs from the majority. These findings highlight a  fundamental risk in GNNs where \textit{underrepresentation of an exogenous characteristic, the graph structure, can cause disparities related to an endongeneous characteristic, the sensitive attribute}, presenting a risk beyond typical fair modeling concerns like sensitive attribute imbalance. Our contributions are outlined below:

\begin{itemize}
\setlength\itemsep{-0.1em}
    \item \textbf{Theory for Local Homophily-Induced Unfairness:} We establish a relationship between fairness and local homophily, showing that node treatment varies with the severity of the OOD problem.
    \item \textbf{New GNN Fairness Benchmarks:} We introduce three benchmarks that feature natural sensitive attributes closely tied to the learning problem and  a wide range of local homophily levels.
    \item \textbf{Semi-synthetic Data Generation:} We propose a re-wiring algorithm to adjust a graph's local homophily distribution using optimal transport between local homophily subgroups.
    \item \textbf{Extensive Empirical Analysis of OOD Problem:} Using our real and semi-synthetic data, we show up to a 24\% increase in unfairness as a byproduct of local homophily. 
\end{itemize}

%% file: PAGES/preliminaries.tex
\section{Preliminaries}

\subsection{Graphs}

Let $G = (V, E, \mathbf{X}, \mathbf{Y})$ denote a simple graph, where $V$ is the node set, $E$ is the edge set, $\mathbf{X} \in \mathbb{R}^{|V| \times f}$ is the feature matrix with $f$ features per node, and $\mathbf{Y} \in \{0, 1\}^{|V| \times c}$ is the label matrix with $c$ classes in one-hot encoded form.
A specific node $i \in G$ has a feature vector $\mathbf{x}_{i}$, a class label $y_{i} \in \{1, \ldots, c\}$, and a one-hot encoded class vector $\mathbf{y}_{i}$. The edge set is often represented using an adjacency matrix, $\mathbf{A} \in \{0, 1\}^{|V| \times |V|}$, where $\mathbf{A}_{i, j} = 1$ indicates an edge between nodes $i$ and $j$. We use $E$ when discussing the edges set of $G$ and $\mathbf{A}$ when referring to matrix computations involving the edges. The \textit{k-hop neighborhood} of node $i \in V$, denoted $N_{k}(i)$, comprises the nodes that can be reached from $i$ within $k$ hops.

\subsection{Node Classification with GNNs}

We focus on the node classification task, where the objective is to learn a mapping from the features $\mathbf{X}$ to the labels $\mathbf{Y}$. In a $k$-layer GNN, this process involves message passing over $k$-hop neighborhoods. The general steps consist of applying a non-linear transformation to the feature matrix $\mathbf{X}$, parameterized by a weight matrix $\mathbf{W}$, and aggregating the features within each node's neighborhood.
The update rule for a node $i$'s representation $\mathbf{r}_{i}$ is given by:
\begin{equation}
\label{eq:gnn_formulation}
    \mathbf{r}_{i}^{l+1} = \text{ENC}(\mathbf{r}_{i}^{l}, \text{AGGR}(\{\mathbf{r}_{v}^{l} \mid v \in N(i) \})), 
\end{equation}
where $l$ is the current layer of the GNN, $\text{AGGR}$ is an aggregation function over the neighboring nodes' representations, and $\text{ENC}$ is an encoding function that synthesizes the representations of node $i$ and its aggregated neighbors.
After $k$ updates, the final class prediction for node $i$ is obtained by taking the $\arg \max$ of its final representation $\mathbf{r}_{i}^{k}$. The initial representations $\mathbf{r}_{i}^{0} = \mathbf{x}_{i}$.

\subsection{Homophily and Heterophily}

We focus on edge homophily and provide the following definitions, noting that there are various ways to express homophily, but many are global and not suited to localized analysis.
First, the global homophily ratio, $h$, describes the overall homophily level of the graph, where $h = 0$ indicates a fully heterophilous graph, and $h = 1$ indicates a fully homophilous graph \cite{mcpherson2001birds}. It is defined as:

\begin{Definition}
\textbf{Global Homophily Ratio.} The global homophily ratio $h$ of a graph's edge set $E$ is the fraction of edges in $E$ that connect nodes $u$ and $v$ with the same label $y_{u}$ and $y_{v}$: 
\begin{equation}
h = \frac{|\{(u,v) \in E \mid y_u = y_v \}|}{|E|}.
\label{eq:global_homophily_ratio}
\end{equation}
\end{Definition}
To analyze homophily at a more granular level, we use the local homophily ratio of a node $t$, $h_{t}$, defined as:

\begin{Definition}
\textbf{Local Homophily Ratio.} \textit{The local homophily ratio $h_{t}$ of a node $t$ is the fraction of edges in its 1-hop neighborhood $N_{1}(t)$ that connect $t$ to a neighbor $u$ with the same class: }
\begin{equation}
h_{t} = \frac{|\{(u,t) : u \in N_{1}(t) \mid y_u = y_{t}\}|}{|N_{1}(t)|}.
\label{eq:local_homophily_ratio}
\end{equation}
\end{Definition}
Since local homophily ratios can vary throughout a graph, we define the local homophily distribution as:

\begin{Definition}
\textbf{Local Homophily Distribution.} The local homophily distribution $P_{G}$ of a graph $G$ is the likelihood of observing a particular local homophily ratio $h_{t}$ within $G$.    
\end{Definition}
We estimate the local homophily distribution by computing the local homophily levels of each node and discretizing these levels into bins to create a probability mass function. We then specify the OOD problem as: 

\begin{Definition}
    \textbf{Out-of-Distribution Homophily Levels.}  {Given a training set of nodes with local homophily distribution $P_{G}$ and a testing set of nodes with local homophily distribution $P_{G'}$, the OOD problem exists when $D(P_{G}, P_{G'}) > \epsilon$, where $D(\cdot, \cdot)$ is a measure of dissimilarity between the train and test set distributions.}
\end{Definition}
When $\epsilon$ is sufficiently large, the train and test nodes are significantly different, constituting an OOD scenario. In our experiments, we use the Earth Mover's Distance (EMD) as $D$ to characterize the distance between the train and test homophily distributions. 

\subsection{Fairness}
In our fairness settings, we assume each node $i$ has a sensitive attribute $s_{i}$, where $s_{i}$ may be  implicitly or explicitly encoded within $\mathbf{x}_{i}$. Under the group fairness paradigm, we use statistical parity (SP) as our notion of fairness with respect to pairs of sensitive attributes, defined as:  
\begin{Definition}
    \textbf{Statistical Parity.} Given two sensitive attributes $s = 0$ and $s = 1$, statistical parity measures the difference in probability of attaining the preferred class $c$ between the two sensitive attributes. 
    \begin{equation}
          |P(y = c | s = 0) - P(y = c | s = 1)|.
    \end{equation}
    
\end{Definition}
For a given model, the probabilities in the SP calculation are estimated empirically using a held-out dataset.

%% file: PAGES/theory.tex
\section{Theoretical Relationship between Local Homophily and Fairness}

Despite independent theoretical studies on either fairness in GNNs \cite{agarwal2021towards, dai2021fairgnn, wang2022fair} or the impact of local homophily on GNNs \cite{mao2023demyst, loveland2023perf}, there are no theoretical analyses to bridge the two concepts. {Thus, we present a theoretical analysis on how nodes with similar labels, but differing sensitive attributes, experience disparate treatment when their \textit{local} homophily levels are underrepresented relative to the training set. We first outline the assumptions and setup for our theoretical framework, then demonstrate how changes in predictions can arise for nodes with differing sensitive attributes.}

\subsection{Theoretical Setup}

{To assess the impact of OOD local homophily levels on GNN predictions, we assume a typical theoretical learning setup used in many previous works \cite{zhu2020h2gcn, loveland2023perf, mao2023demyst, wang2022fair, du2022gbk}.} Specifically, we focus on a binary classification task with binary sensitive attributes and consider a GNN where the representations $\mathbf{r}_i$ for a node $i$, with degree $d$ and local homophily level $h_i$, are computed using an aggregation function $\text{AGGR}(i) = \sum_{v \in N(i)} \mathbf{r}_v$ and an update function $\text{ENC}(i) = (\mathbf{r}_i + \text{AGGR}(i))\mathbf{W}$. In matrix form, the GNN can be represented as $\mathbf{(A+I)XW}$.

A node $i$'s features are designed to encode both the label $y_{i}$ and sensitive attribute $s_{i}$. 
Specifically, $i$'s feature vector is the concatenation of the label and sensitive attribute features, $\mathbf{x}_{i} = [x_{i}^{(l)} \mid\mid  x_{i}^{(s)} ]$, where 
(1) the label is encoded as $x_{i}^{(l)} = -p_i$ if $y_i = 0$, and $x_{i}^{(l)} = p_i$ if $y_i = 1$, with $p_i \overset{\mathrm{i.i.d}}\sim  \mathcal{N}(\mu_{l}, \sigma_{l})$; 
(2) the sensitive attribute is encoded as $x_{i}^{(s)} = -q_i$ when  $s_i=0$ and  $x_{i}^{(s)} = q_i$ when $s_{i} = 1$, with $q_i \overset{\mathrm{i.i.d}}\sim  \mathcal{N}(\mu_{s}, \sigma_{s})$. 
When $\mu_{l}=$ $0$, the variable $x_{i}^{(l)}$ follows the same distribution when $y_{i}=$  $0$ or $1$. In contrast, a larger $\mu_{l}$ encourages the features to deviate and provide discriminability on the label. Similar logic holds for $\mu_{s}$ and $x_{i}^{(s)}$.

\subsection{Impact of Out-of-Distribution Problem on Fairness}

We explicitly study a biased learning setup where a GNN is trained on $k$ data points where $y = s = 0$ and $n - k$ data points where $y = s = 1$, creating a positive association between the class and sensitive attribute. After solving for the expected weight matrix $\mathbb{E}[\mathbf{W}]$, we attain predictions for two test nodes $u$ and $v$, with the same label (WLOG let $y_{u} = y_{v} = 0$), but differing sensitive attributes. We also assume that $u$ and $v$ have the same local homophily $h + \alpha$, where $\alpha$ is the relative shift in local homophily level from the global homophily level $h$. We show that as $\alpha$ increases in magnitude, denoting increase in OOD distance, the gap in expected predictions of the two test nodes also increases.

\begin{theorem}
Consider test nodes $u$ and $v$ with local homophily ratios $h + \alpha$, labels $y_u = 0$ and $y_v = 0$, and sensitive attributes $s_u = 0$ and $s_v = 1$. The difference in their expected logit associated with the correct label, $\mathbb{E}[\mathbf{p}_{u}]_{y_{u}}$ and $\mathbb{E}[\mathbf{p}_{v}]_{y_{v}}$, is:
\begin{equation}
     \mathbb{E}[\mathbf{p}_{u}]_{y_{u}} - \mathbb{E}[\mathbf{p}_{v}]_{y_{v}} =  
            \frac{\mu_{s}^{2}k(1+d(2h + 2\alpha - 1))}{n(1 + d(2h-1))((\mu_{l}^2 + \mu_{s}^2)}.
\end{equation}

\label{theorem:diff_in_sens_logit}
\end{theorem}

The proof for this theorem is provided in App.~\ref{proof}. 
A large difference, either positive or negative, increases the likelihood of differing predictions for nodes $u$ and $v$, suggesting disparate treatment of individuals with different sensitive attributes. 
To further assess the criteria that disparate treatment arises, we must determine when $|\mathbb{E}[\mathbf{p}_{u}]_{y_{u}} - \mathbb{E}[\mathbf{p}_{v}]_{y_{v}}| > 0$, i.e. there is preferred treatment towards a sensitive attribute. This inequality holds when $\mathbb{E}[\mathbf{p}_{u}]_{y_{u}} - \mathbb{E}[\mathbf{p}_{v}]_{y_{v}} \ne 0$. Assuming $\mu_{s} \ne 0$ and $k > 0$, $\mathbb{E}[\mathbf{p}_{u}]_{y_{u}} - \mathbb{E}[\mathbf{p}_{v}]_{y_{v}} \ne 0$ when $\alpha \ne 0$. These results indicate that drifting away from the global homophily level can increase the distance between predicted logits. 
If $c = 0$ in our SP definition, our theorem can also be interpreted as characterizing the average increase in SP for different $\alpha$ values. In the following sections, we validate our findings on semi-synthetic and real-world datasets.

%% file: PAGES/datasets_and_synth_method.tex
\section{New Real World Datasets}

\label{datasets}
In this section, we outline current challenges with fairness benchmarks in node classification and present our improved datasets. Our datasets possess diverse homophily patterns while ensuring reproducibility and well-motivated fairness implications. These datasets are also necessary to study fair GNNs for node classification given most fairness datasets are tailored for link prediction \cite{chen2024fairness}. 
Through these real-world datasets, we expand the scenarios for studying GNNs, thereby minimizing unexpected behavior upon deployment.

\subsection{Limitations of Current Datasets}

While the number of GNNs continues to expand, the datasets to study GNN fairness remain limited \cite{dong2023fairness}. Moreover, existing datasets exhibit fundamental issues, preventing systematic evaluation of how local homophily impacts fairness. 
In this section, we provide a high-level overview of the identified failure modes of current GNN fairness datasets, with thorough discussion and example datasets of each failure mode provided in App. \ref{prev_datasets}.

First, we focus on diversity with respect to homophily patterns. Given prominent fairness datasets are built around similarity networks \cite{agarwal2021towards, dong2023elegant, chen2024fairness}, their respective distributions tend to be homophilous, making it difficult to study diverse OOD settings. Second, many fairness datasets are ambiguous given they have multiple variants with missing information \cite{dong2023fairness, dai2021fairgnn, chen2024fairness}. Finally, a significant number of fairness benchmarks are datasets without sensitive attributes or fairness contexts\cite{chen2024fairness}. Instead, structural attributes are arbitrarily chosen as sensitive attributes despite not being representative of disparate treatment \cite{zhao2022towards, dong2023fairness}. Together, these three challenges hinder the progression of GNN fairness research, particularly as it pertains to the OOD problem.

\begin{figure}[ht]
    \centering
    \includegraphics[width=8.5cm]{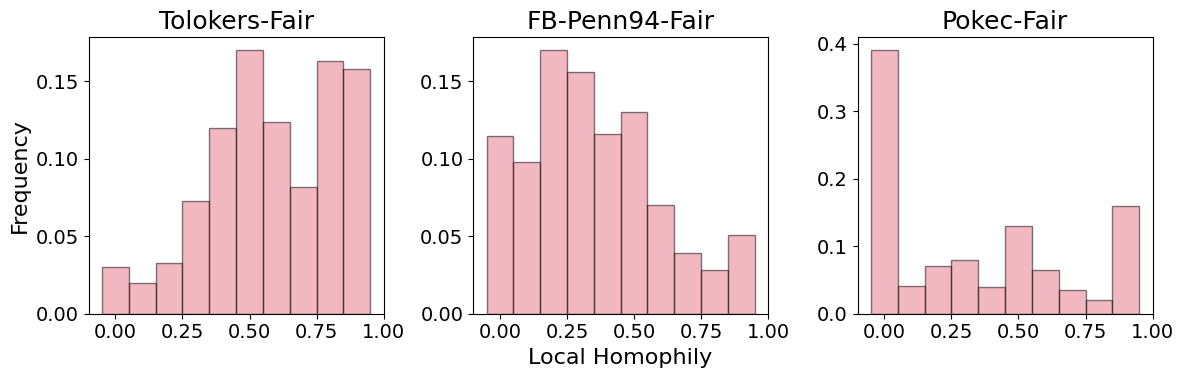}
    \caption{\textbf{Local Homophily Distributions for Proposed Datasets.} Each dataset displays a wide array of local and global homophily levels. }
    \label{fig:local_homophily}
\end{figure}

\subsection{Proposed Real-World Datasets}

The proposed datasets are fully labeled with both class and sensitive attribute information. We focus on datasets with diverse local homophily patterns to study different OOD scenarios. To mitigate remote nodes, we extract the largest connected component from each dataset which satisfies the above criteria. Stats are given in Table \ref{table:dataset_stats}, and the homophily distributions are given in Figure \ref{fig:local_homophily}. 

\noindent \textbf{Tolokers-Fair:} 
The Tolokers graph, derived from users of the Tolokers crowdsourcing platform, was introduced by Lim et al. \cite{platonov2022critical} as a large-scale benchmark for GNNs. We introduce a fairness-based learning task: predicting whether an individual contributor should be banned from the platform, using dominant language (English vs. non-English) as the sensitive attribute. This task is motivated by potential language biases. The features are retained based on their original implementation. Tolokers-Fair is skewed homophilous ($h = 0.58$).

\noindent \textbf{FB-Penn94-Fair:} 
The FB graphs, based on Facebook social networks, were introduced by Leskovec et al. \cite{leskovec2012fb}. Despite their use as benchmarks \cite{lim2022hetero}, the FB datasets have not been directly used for fairness. To study fairness, we propose predicting the major of a student (instead of gender in previous studies) and using gender as a sensitive attribute, addressing gender inequity in college majors. As the major has high cardinality, we extract the five most common majors and subsample nodes with these majors and a gender attribute. The largest component is taken as the dataset. The features are retained based on their original implementation. FB-Penn94-Fair is heterophilous ($h = 0.39$).

\noindent \textbf{Pokec-Fair:} 
Originally proposed by \cite{dai2021fairgnn}, Pokec is a fairness dataset with artificially scrubbed subgraphs retaining only 1\% of the labels and sensitive attributes. No information is provided to recover the scrubbed information, nor the subgraphs. To perform local analyses, we create Pokek-Fair with the task of predicting occupation while using gender as the sensitive attribute. We do not perform major region sampling as done in \cite{dai2021fairgnn}, given region is a high cardinality feature with many missing values. Similar to FB-Penn94-Fair, we extract the 5-most common jobs, excluding unemployment, leaving classes ``service work", ``construction", ``finance", ``healthcare", and ``transport", and subsample the users that work in these fields and possess a gender attribute. As noted in \cite{lim2022hetero}, the pre-processing of features is highly variable, with \cite{dai2021fairgnn} reporting 59 features, and \cite{lim2022hetero} reporting 65 features. As open source access to the feature processing is not provided, we reduce inconsistencies by removing the free text features, and only retain the categorical and numerical features. The Pokec-Fair dataset is heterophilous ($h = 0.38$).

\begin{table}[]
\caption{Proposed Dataset Statistics and Details}
\centering
\footnotesize
\begin{tabular}{@{}lrrll@{}}
\toprule
Dataset     & Nodes & Edges  & Task             & \shortstack{Sens. \\ Attr.} \\ \midrule
Tolokers-Fair    & 11,758 & 519K & Banned(2) & Language \\
FB-Penn94-Fair   & 7,016  & 59,845  & Major(5)  & Gender   \\ 
Pokec-Fair       & 69,949 & 130K & Job(5)    & Gender   \\
\bottomrule
\end{tabular}
\vspace{-.3cm}
\label{table:dataset_stats}
\end{table}

\section{Semi-Synthetic Data Generation}

In addition to our proposed real-world datasets, we introduce a semi-synthetic graph generator that enables the exploration of diverse local homophily distributions.
Semi-synthetic data generation is commonly used to assess GNN behavior under varied homophily conditions~\cite{karimi2018homophily, zhu2020h2gcn, mao2023demyst}, but existing methods only control global homophily and often inflate node degrees by adding edges. To address these limitations, we propose a strategy that manages homophily at the node level without requiring a full node-class interaction matrix~\cite{maekawa2023gencat}. At a high level, we achieve this by defining a goal homophily distribution, parameterized by the beta probability distribution, and re-wiring a small set of edges around select nodes. This minimally disrupts the original graph, adjusting only necessary nodes to match the goal distribution. Although we do not explicitly consider feature information during re-wiring, we adhere to the common assumption that features encode the label information, allowing the underlying signal to be extracted during message passing. We describe our method below. 

\begin{figure}[ht!] \centering \subfloat{ \includegraphics[width=0.365\textwidth]{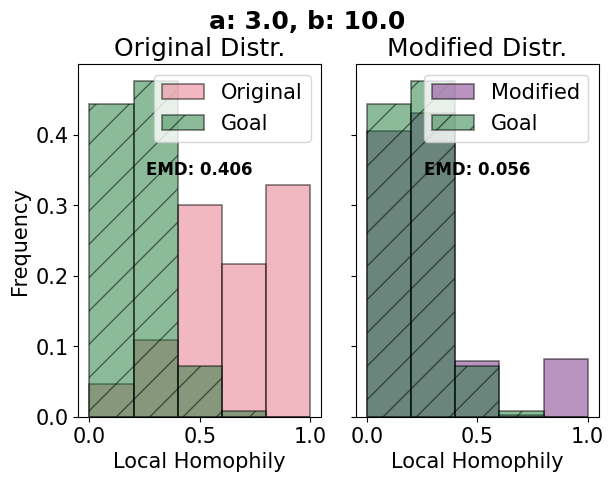} }\\ \subfloat{ \includegraphics[width=0.365\textwidth]{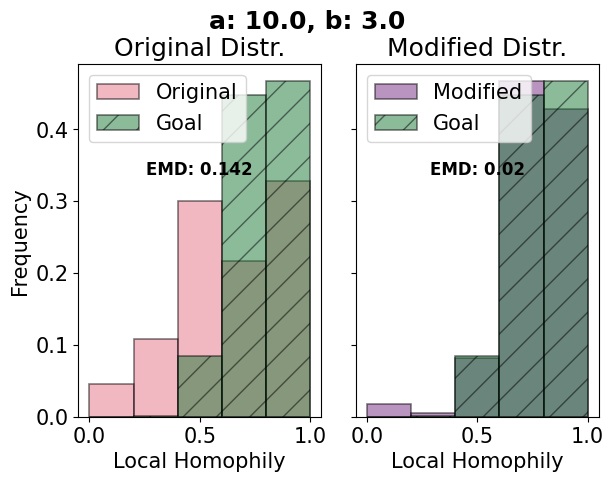} } \caption{\textbf{Comparison of Original, Generated, and Goal Distributions for Tolokers-Fair Dataset.} Red and purple (no hatching) denote the original and modified distributions, respectively. The green (hatching) denotes the goal distribution. Overlap between the distributions indicates better re-wiring.} \label{fig:distr_gen} \end{figure}

\vspace{0.1cm}
\noindent \textbf{Parameterizing Goal Homophily Distribution: } We parameterize the goal homophily distribution as a beta distribution, where $X \sim Beta(\alpha, \beta)$ with probability density function $C(x)^{\alpha-1}(1-x)^{\beta-1}$, and normalization constant $C$. Specifying $\alpha$ and $\beta$ determines the likelihood of sampling a particular local homophily ratio, as dictated by $\mathbb{E}[X] = \dfrac{\alpha}{\alpha + \beta}$. Thus, when $\alpha > \beta$, the distribution will skew homophilous, while $\alpha < \beta$ will skew heterophilous. The spread of the distribution is additionally controlled by $\text{Var}[X] = \dfrac{\alpha \beta}{(\alpha + \beta)^{2}(\alpha + \beta + 1)}$, where larger values create well peaked distributions.

\vspace{0.1cm}
\noindent \textbf{Determining Edges to Re-wire: }
To achieve the goal homophily distribution for a graph, we start by specifying the target homophily distribution $Q$ (coming from $Beta(\alpha, \beta)$), and computing the empirical homophily histogram $P_{G}$ for graph $G$ with $b$ bins. We solve an optimal transport problem to align $P_{G}$ with $Q$, yielding a transportation matrix $\mathbf{T} \in [0, 1]^{b \times b}$. 
Each matrix entry $\mathbf{T}_{i, j}$ denotes the proportion of nodes whose local homophily ratio falls within bin $i$ and needs to be modified to fall within bin $j$. Nodes that already meet the goal homophily level are not modified and retain their original structure. 
For the other nodes, we precompute the number of edges that need to be modified to reach their goal homophily level. For a node $i$ with local homophily $h_{i}$, goal homophily $h_{g}$, and degree $d_{i}$, the total number of edges to move, $e_{i}$, can be bounded as $|(h_g - h_i)d_{i}| \le e_{i} \le |\dfrac{(h_{g} - h_{i})d}{1-h_{g}}|$ when $h_{i} < h_{g}$ and $|(h_g - h_i)d_{i}| \le e_{i} \le |\dfrac{(h_{i} - h_{g})d}{h_{g}}|$ when $h_{i} > h_{g}$. 
The lower and upper bounds denote the number of edges that are moved when purely re-wiring  (no degree change) or adding edges (degree change). 

\vspace{0.1cm}
\noindent \textbf{Our Minimally Disruptive Re-wiring Model: } Our generation (pseudocode in Appendix \ref{semisynth_psuedo}) begins by sampling nodes from $G$ based on the matrix $\mathbf{T}$. These are denoted as source nodes. The number of edges to be modified in each source node's neighborhood is determined using the lower bounds above, with the criteria that removing an edge optimizes both the source and neighboring nodes homophily levels. Then, edges removed from the source node are re-wired to candidate nodes outside the neighborhood, again optimizing both the source and candidate nodes' local homophily.
As an example, if a node is to become more homophilous, a heterophilous edge will be removed from a neighboring node that also needs to become more homophilous. Then, the edge is converted into a homophilous edge between a candidate node that also needs to become more homophilous.
If re-wiring alters a neighboring or candidate node's degree, the lower bound is recalculated. 

As re-wiring is required to strictly decrease the distance between the current and goal homophily level for any nodes, it may not be possible to attain the goal homophily ratio under such constraint. Thus, we provide a secondary refinement phase where edges are added to further adjust the local homophily ratio of the node, at the expense of modify the degree distribution slightly. Specifically, the upper bounds above are computed for each node, and edges are added between source and candidate nodes to mutually benefit their local homophily level. This process continues until all nodes achieve their targets or no further beneficial modifications are possible. Note that this process can accommodate both binary and multi-class settings. 

\vspace{0.1cm}
\noindent \textbf{Evaluation:} We assess the re-wiring quality using the EMD between the original/goal and generated/goal distributions. As seen in Table \ref{table:synth_eval} in the Appendix, each base dataset experiences significant EMD reductions for generated distributions, indicating successful re-wiring. 
To further elucidate the re-wiring solution in relation to the EMD value, we provide distributions for the Tolokers-Fair dataset in Figure \ref{fig:distr_gen}. For each case, the modified distributions match well with the goal distributions, demonstrating that the semi-synthetic generator can act as a promising tool to explore the space of local homophily distributions. Through these capabilities, GNN fairness can be comprehensively assessed under a variety of connectivity patterns beyond what is available in benchmarks.

%% file: PAGES/results.tex
\section{Empirical Analysis}
This section studies the relationship between OOD local homophily levels and unfairness through empirical analysis. We leverage the proposed real-world and semi-synthetic datasets and focus on two fundamental questions: (RQ1) \textit{How does the global and local homophily levels of a graph impact fairness?}, and (RQ2) \textit{How does GNN design mitigate unfairness in the OOD setting?}

\subsection{Train/Val/Test Processing} 
\begin{figure}[h!]
    
    \centering
    \includegraphics[width=7.6cm]{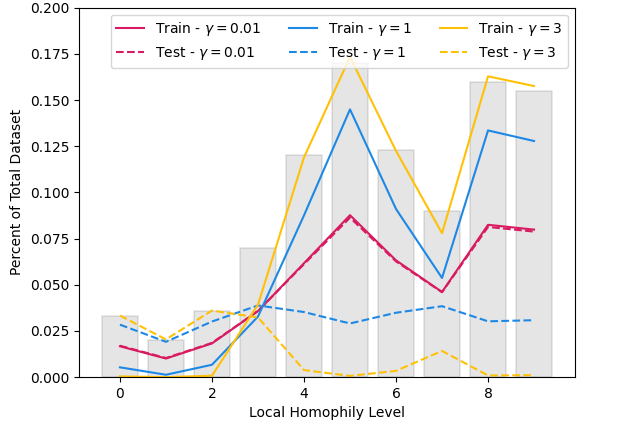}
    \vspace{-0.2cm}
    \caption{\textbf{Example Train/Test Splits with differing $\boldsymbol{\gamma}$.} Gray bars denote the $P_{G}$ for the Tolokers-Fair dataset. As $\gamma$ increases, the train and test sets become more disjoint, leading to more OOD samples in the test set. When $\gamma \approx 0$, train and test are identical.}
    \label{fig:ood_train_test}
    \vspace{-0.2cm}
\end{figure}

To study how the OOD problem impacts fairness, we create two test sets, one in-distribution and one OOD. Given a dataset's local homophily distribution $P_{G}$, we generate these two scenarios by concentrating the mass of $P_{G}$ and sampling from the new distribution. Specifically, each bin $b \in B$ is raised to the $\gamma$ power, and then normalized, resulting in $P_{G}^{\gamma}$. We then compute an inverted distribution $\bar{P}_{G}^{\gamma}$, where $\bar{P}_{G,b}^{\gamma} = (P_{G,b}^{\gamma})^{-1} / \sum_{b \in B} (P_{G,b}^{\gamma})^{-1}$. Parameter $\gamma \in [0, \infty)$ allows us to choose the concentration of mass within ${P}_{G, b}^{\gamma}$ and $\bar{P}_{G, b}^{\gamma}$, changing the OOD severity. When $\gamma = 0$, the test set is in-distribution, while when $\gamma = 3$ the test set is OOD.
The proportion of training and testing nodes in each homophily range bin $b$ is determined by $P_{G,b}^{\gamma} / (P_{G, b}^{\gamma} + \bar{P}_{G, b}^{\gamma})$ and $\bar{P}_{G,b}^{\gamma} / (P_{G, b}^{\gamma} + \bar{P}_{G, b}^{\gamma})$, respectively. In both $\gamma = 0$ ( standard splitting protocol), and $\gamma = 3$ (our stratified protocol), the train and test set are randomly sampled with an 80/20 split. Then, the validation set is sampled as a subset from the training set with an 80/20 train/val ratio. Figure \ref{fig:ood_train_test} shows splits for different $\gamma$ values.

\begin{table*}[h!] \scriptsize \centering \caption{Model Performance Across Real World Datasets. The data includes F1 and statistical parity (SP) for \(\gamma = 0\) and \(\gamma = 3\), along with changes in F1 and SP.}\begin{tabular}{@{}ccccccccccc@{}} \toprule \textbf{Dataset} & \textbf{Model} & \multicolumn{2}{c}{$\gamma = 0$} & \multicolumn{2}{c}{$\gamma = 3$} & \multicolumn{2}{c}{$\Delta$} & \textbf{Rank} \\ \cmidrule(lr){3-4} \cmidrule(lr){5-6} \cmidrule(lr){7-8} & & \textbf{F1} & \textbf{SP} & \textbf{F1} & \textbf{SP} & \cellcolor{gray!30}\textbf{F1} & \cellcolor{gray!30}\textbf{SP} & \textbf{} \\ \midrule
\multirow{7}{*}{\textbf{Tolokers-Fair}} 
 & \textbf{GCN} & 0.06 (0.02) & -0.02 (0.02) & -0.03 (0.02) & 0.07 (0.02) & \cellcolor{gray!30}-0.09 & \cellcolor{gray!30}0.09 & 3 \\
 & \textbf{SAGE} & 0.10 (0.06) & -0.05 (0.01) & -0.11 (0.02) & -0.03 (0.01) & \cellcolor{gray!30}-0.21 & \cellcolor{gray!30}0.02 & 2 \\
 & \textbf{LINKX} & 0.08 (0.05) & 0.02 (0.03) & -0.12 (0.04) & 0.01 (0.02) & \cellcolor{gray!30}-0.20 & \cellcolor{gray!30}-0.01 & 1 \\
 & \textbf{FairGCN} & 0.15 (0.04) & -0.17 (0.01) & -0.03 (0.02) & -0.06 (0.01) & \cellcolor{gray!30}-0.18 & \cellcolor{gray!30}0.11 & 4 \\
 & \textbf{FairSAGE} & 0.16 (0.04) & -0.19 (0.01) & -0.03 (0.02) & -0.07 (0.01) & \cellcolor{gray!30}-0.19 & \cellcolor{gray!30}0.12 & 5 \\
 & \textbf{NiftyGCN} & -0.63 (0.04) & -0.20 (0.01) & -0.59 (0.02) & -0.07 (0.01) & \cellcolor{gray!30}0.04 & \cellcolor{gray!30}0.13 & 6 \\
 & \textbf{NiftySAGE} & -0.62 (0.04) & -0.20 (0.01) & -0.56 (0.02) & -0.06 (0.01) & \cellcolor{gray!30}0.06 & \cellcolor{gray!30}0.14 & 7 \\ \midrule

\multirow{7}{*}{\textbf{Pokec-Fair}}
 & \textbf{GCN} & 0.02 (0.01) & -0.34 (0.02) & 0.02 (0.02) & -0.22 (0.03) & \cellcolor{gray!30}0.0 & \cellcolor{gray!30}0.12 & 4 \\
 & \textbf{SAGE} & 0.04 (0.01) & -0.15 (0.03) & 0.02 (0.03) & 0.08 (0.03) & \cellcolor{gray!30}-0.02 & \cellcolor{gray!30}0.23 & 6 \\
 & \textbf{LINKX} & -0.02 (0.02) & -0.41 (0.0) & -0.08 (0.03) & -0.35 (0.03) & \cellcolor{gray!30}-0.06 & \cellcolor{gray!30}0.06 & 1 \\
 & \textbf{FairGCN} & -0.01 (0.01) & -0.71 (0.02) & 0.08 (0.02) & -0.54 (0.04) & \cellcolor{gray!30}0.09 & \cellcolor{gray!30}0.17 & 5 \\
 & \textbf{FairSAGE} & -0.01 (0.02) & -0.70 (0.02) & 0.06 (0.02) & -0.46 (0.06) & \cellcolor{gray!30}0.07 & \cellcolor{gray!30}0.24 & 3 \\
 & \textbf{NiftyGCN} & -0.02 (0.01) & -0.76 (0.01) & 0.06 (0.02) & -0.70 (0.03) & \cellcolor{gray!30}0.08 & \cellcolor{gray!30}0.06 & 1 \\
 & \textbf{NiftySAGE} & -0.02 (0.01) & -0.73 (0.01) & 0.05 (0.01) & -0.66 (0.02) & \cellcolor{gray!30}0.07 & \cellcolor{gray!30}0.07 & 3 \\ \midrule

\multirow{7}{*}{\textbf{FB-Penn94-Fair}}
 & \textbf{GCN} & 0.02 (0.01) & 0.06 (0.01) & 0.0 (0.01) & 0.01 (0.01) & \cellcolor{gray!30}-0.02 & \cellcolor{gray!30}-0.05 & 3 \\
 & \textbf{SAGE} & 0.07 (0.01) & 0.05 (0.01) & 0.07 (0.01) & 0.02 (0.01) & \cellcolor{gray!30}0.0 & \cellcolor{gray!30}-0.03 & 5 \\
 & \textbf{LINKX} & 0.08 (0.01) & 0.10 (0.03) & 0.07 (0.02) & 0.04 (0.01) & \cellcolor{gray!30}-0.01 & \cellcolor{gray!30}-0.06 & 1 \\
 & \textbf{FairGCN} & 0.01 (0.01) & 0.03 (0.0) & 0.02 (0.02) & 0.01 (0.02) & \cellcolor{gray!30}0.01 & \cellcolor{gray!30}-0.02 & 6 \\
 & \textbf{FairSAGE} & 0.0 (0.01) & 0.07 (0.01) & -0.05 (0.02) & 0.01 (0.01) & \cellcolor{gray!30}-0.05 & \cellcolor{gray!30}-0.06 & 1 \\
 & \textbf{NiftyGCN} & 0.0 (0.04) & 0.04 (0.02) & 0.01 (0.05) & 0.0 (0.01) & \cellcolor{gray!30}0.01 & \cellcolor{gray!30}-0.04 & 4 \\
 & \textbf{NiftySAGE} & 0.03 (0.01) & 0.03 (0.01) & -0.01 (0.0) & 0.05 (0.01) & \cellcolor{gray!30}-0.04 & \cellcolor{gray!30}0.02 & 7 \\
\bottomrule

\end{tabular}
\label{table:results_grouped}
\end{table*}

\subsection{Models and Evaluation}

To understand how (un)fairness can arise, we evaluate three classes of GNNs: (1) Classic GNNs, (2) GNNs designed to accommodate homophily and heterophily, and (3) GNNs built with fairness considerations. Our goal is to address whether local homophily induced fairness degradation can be addressed by heterophilous learning mechanisms. We use GCN \cite{kipf2016semi} to represent our classic GNN, LINKX \cite{lim2022hetero} as our heterophilous model, and GraphSAGE \cite{hamilton2017inductive} as a midpoint between the two. Specifically, GCN offers a simple degree weighted aggregation, while GraphSAGE and LINKX both de-couple the encoding of the ego and neighbor embeddings -- a key design shown to help learn over heterophily. Additionally, LINKX seperates the structure and feature learning to further help in heterophilous settings. 
For fairness-specific GNNs, we employ Nifty \cite{agarwal2021towards} and FairGNN \cite{dai2021fairgnn}, each with GCN and GraphSAGE backbones. An MLP is trained as a graph-agnostic baseline.

Hyperparameters are chosen via cross-validation. Each experiment is repeated three times. Performance and fairness are measured using the micro-F1 score and SP, respectively. As the common formulation of SP relies on binary properties \cite{agarwal18fairdp, agarwal2021towards}, we also consider a multi-class variant which computes the maximum SP for all pairs of classes \cite{denis2021fairness, rouzot2022learning}. 
For the real-world results in Table \ref{table:results_grouped}, we provide F1 and SP for both $\gamma$ values, while for the semi-synthetic results in Table \ref{table:synth_results_table} we provide the  change in F1 and SP between $\gamma = 3$ and $\gamma = 0$. 
For all results, the MLP F1 and SP scores are subtracted from GNN scores to control for spurious feature relationships, helping to isolate the impact of homophily.

\subsection{Results}

\begin{table*}[h]
\scriptsize
\centering
\caption{Model Performance Across Semi-Synthetic Datasets with Parameter Settings $\alpha=3.0, \beta=10.0$ (heterophilous) and $\alpha=10.0, \beta=3.0$ (homophilous). The data includes changes in F1 and SP between $\gamma = 0$ and $\gamma = 3$.}
\begin{tabularx}{\textwidth}{@{}l*{3}{*{4}{>{\centering\arraybackslash}X}}@{}}
\toprule
 & \multicolumn{4}{c}{\textbf{Tolokers-Fair Semi-Synth}} & \multicolumn{4}{c}{\textbf{Pokec-Fair Semi-Synth}} & \multicolumn{4}{c}{\textbf{FB-Penn94-Fair Semi-Synth}} \\ \cmidrule(lr){2-5} \cmidrule(lr){6-9} \cmidrule(lr){10-13}
\textbf{Model} & \multicolumn{2}{c}{\(\alpha = 3.0, \beta = 10.0\)} & \multicolumn{2}{c}{\(\alpha = 10.0, \beta = 3.0\)} & \multicolumn{2}{c}{\(\alpha = 3.0, \beta = 10.0\)} & \multicolumn{2}{c}{\(\alpha = 10.0, \beta = 3.0\)} & \multicolumn{2}{c}{\(\alpha = 3.0, \beta = 10.0\)} & \multicolumn{2}{c}{\(\alpha = 10.0, \beta = 3.0\)} \\ \cmidrule(lr){2-3} \cmidrule(lr){4-5} \cmidrule(lr){6-7} \cmidrule(lr){8-9} \cmidrule(lr){10-11} \cmidrule(lr){12-13}
 & \(\Delta \text{F1}\) & \(\Delta \text{SP}\) & \(\Delta \text{F1}\) & \(\Delta \text{SP}\) & \(\Delta \text{F1}\) & \(\Delta \text{SP}\) & \(\Delta \text{F1}\) & \(\Delta \text{SP}\) & \(\Delta \text{F1}\) & \(\Delta \text{SP}\) & \(\Delta \text{F1}\) & \(\Delta \text{SP}\) \\ \midrule
\textbf{GCN} & -0.13 & \cellcolor{gray!30}0.09 & -0.21 & \cellcolor{gray!30}0.08 & 0.02 & \cellcolor{gray!30}-0.05 & -0.04 & \cellcolor{gray!30}0.08 & -0.01 & \cellcolor{gray!30}0.12 & -0.09 & \cellcolor{gray!30}0.06 \\ 
\textbf{SAGE} & -0.16 & \cellcolor{gray!30}-0.09 & -0.28 & \cellcolor{gray!30}0.11 & -0.01 & \cellcolor{gray!30}-0.07 & -0.12 & \cellcolor{gray!30}0.10 & -0.01 & \cellcolor{gray!30}0.14 & -0.35 & \cellcolor{gray!30}0.27 \\ 
\textbf{LINKX} & -0.12 & \cellcolor{gray!30}-0.09 & -0.20 & \cellcolor{gray!30}0.07 & 0.0 & \cellcolor{gray!30}-0.03 & -0.04 & \cellcolor{gray!30}0.22 & 0.03 & \cellcolor{gray!30}-0.08 & -0.21 & \cellcolor{gray!30}0.22 \\ 
\textbf{FairGCN} & -0.01 & \cellcolor{gray!30}-0.14 & -0.18 & \cellcolor{gray!30}0.11 & 0.05 & \cellcolor{gray!30}-0.15 & -0.03 & \cellcolor{gray!30}-0.06 & 0.01 & \cellcolor{gray!30}-0.01 & -0.03 & \cellcolor{gray!30}0.05 \\ 
\textbf{FairSAGE} & 0.02 & \cellcolor{gray!30}-0.10 & -0.18 & \cellcolor{gray!30}0.10 & 0.05 & \cellcolor{gray!30}-0.12 & -0.02 & \cellcolor{gray!30}-0.04 & -0.02 & \cellcolor{gray!30}-0.11 & 0.02 & \cellcolor{gray!30}-0.10 \\ 
\textbf{NiftyGCN} & -0.05 & \cellcolor{gray!30}-0.14 & -0.08 & \cellcolor{gray!30}0.11 & 0.05 & \cellcolor{gray!30}-0.16 & 0.0 & \cellcolor{gray!30}0.05 & 0.05 & \cellcolor{gray!30}0.06 & -0.04 & \cellcolor{gray!30}0.15 \\ 
\textbf{NiftySAGE} & -0.32 & \cellcolor{gray!30}-0.15 & -0.06 & \cellcolor{gray!30}0.11 & 0.05 & \cellcolor{gray!30}-0.16 & 0.0 & \cellcolor{gray!30}0.30 & -0.03 & \cellcolor{gray!30}0.10 & -0.05 & \cellcolor{gray!30}-0.14 \\ 
\bottomrule
\end{tabularx}
\label{table:synth_results_table}
\end{table*}

In this section, we present the results of our empirical analysis, revealing insights into how homophily can induce unfairness and the effectiveness of different GNN designs in mitigating this. We begin with a high-level analysis to establish that the OOD problem exists, and then move to granular dataset- and model-level analysis to explain drivers for unfairness. 

\vspace{0.1cm}
\noindent \textbf{(RQ1) The Impact of Homophily on Fairness:} 
Table \ref{table:results_grouped} summarizes the performance and fairness across different datasets. On average, we observe a 6.3\% increase in SP, indicating increased unfairness. Notably, the FB-Penn94-Fair dataset exhibits a 3.4\% average decrease in SP, suggesting it is generally unaffected by the OOD problem. In contrast, the Tolokers-Fair and Pokec-Fair datasets show increases in SP of 8.6\% and 13.6\%, respectively.
When we quantify the average EMD between the train and test sets for each dataset, we find the following values: 0.037 for FB-Penn94-Fair, 0.082 for Tolokers-Fair, and 0.179 for Pokec-Fair,  \textit{elucidating a near-linear relationship between OOD severity and unfairness}. This additionally corresponds with our theoretical analysis which relates the homophily shift level $\alpha$ to unfairness. 
We further characterize the impact of global homophily through our semi-synthetic setting where EMD values remain roughly constant. This allows us to isolate changes in SP as a factor of homophily level, rather than EMD. We find that globally homophilous datasets tend to induce unfairness on heterophilous nodes, increasing average SP by 8.8\%. Conversely, globally heterophilous datasets tend to improve fairness for homophilous nodes, reducing average SP by 5.4\%. This pattern agrees and further explains FB-Penn94-Fair's fairness levels, given it is global heterophilous with low OOD severity.

Overall, these results suggest a nuanced interplay between global and local homophily levels. Specifically, globally heterophilous datasets are beneficial for improving fairness in OOD settings until the OOD data points become significantly far from the training set. This finding implies that both local and global homophily levels must be considered together to fully understand fairness patterns. Previous works have addressed the challenge of fairness when applying GNNs to globally homophilous graphs but have overlooked (a)~the local homophily level and (b)~the shift of local homophily level relative to the global level \cite{spinelli2021dropout}.

\begin{figure}[h!]
    \centering
    \includegraphics[width=7.25cm]{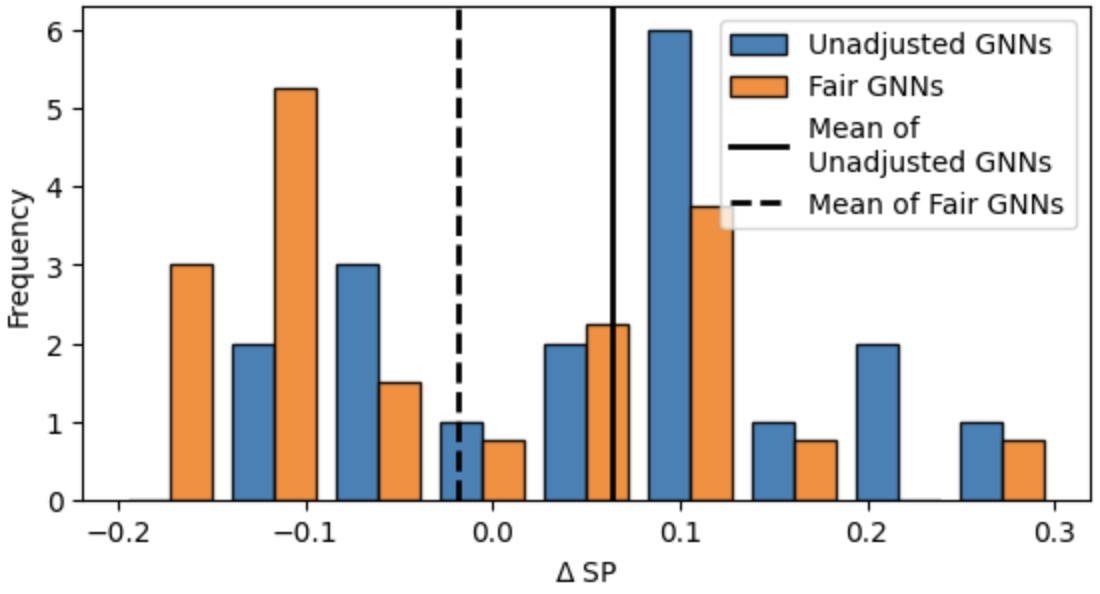}
    \caption{\textbf{Distribution of $\Delta$SP for Real and Semi-Synthetic Datasets.} The histogram indicates that unadjusted GNNs (GCN, SAGE, and LINKX) produce more frequent SP increases compared to fairness-adjusted GNNs (Nifty and FairGNN).} 
    
    \label{fig:ood_dist}
\end{figure}

\noindent \textbf{(RQ2) The Impact of GNN Design on Fairness:} We now analyze different GNN designs as outlined in the experimental setup. As opposed to RQ1, we look at two properties, the initial SP when $\gamma = 0$, as well as $\Delta$SP. To ease comparisons, in Table $\ref{table:results_grouped}$ each model is ranked across datasets, with lower rank indicating less of a change in SP. Interestingly, LINKX is ranked among the best across each dataset, often producing smaller drops in fairness as opposed to the fair GNN architectures. Note that across each dataset the initial $SP$ when $\gamma = 0$ is significantly lower for these fair models, highlighting \textit{fair GNNs are beneficial for in-distribution data points, but are more susceptible to OOD fairness degradation}. This is further seen in the semi-synthetic data as when the OOD severity is controlled through similar homophily distributions, the GNNs without fairness adjustment do significantly worse on average. The comparison between unadjusted and fairness-adjusted GNNs is provided in \ref{fig:ood_dist} across both the real and semi-synthetic datasets, demonstrating an $8.2\%$ difference in average $\Delta$SP.  

%% file: PAGES/related_work.tex
\section{Related Work}

While the majority of our related work is introduced in Sections \ref{intro} and \ref{datasets}, here we provide previous fairness methods not specifically related to homophily. 
Many works on GNN fairness aim to ensure that the representations learned by GNNs do not heavily rely on sensitive attributes \cite{chen2024fairness, dong2023fairness}. There are two common paradigms: (1) pre-processing, where the original graph is altered to promote fairness, and (2) in-processing, where the computation graph is adjusted to facilitate fair representations.
FairWalk \cite{rahman2019fairwalk} proposes a random walk technique that ensures nodes from different sensitive groups are sampled at equal rates. Similarly, FairAdj \cite{li2020dyadic} and FairDrop \cite{spinelli2021dropout} modify the adjacency matrix to create more balanced neighborhoods, reducing the influence of sensitive attributes.
For in-processing, adversarial debiasing has been utilized to mitigate the GNN's reliance on sensitive attributes \cite{dai2021fairgnn}. Nifty \cite{agarwal2021towards} approaches the problem from a representation perspective by promoting layer-wise re-normalization and neighborhood augmentation. Re-weighting has also inspired methods like FairGAT and FairVGNN, which tend to work well for in-distribution nodes \cite{kose2023fairgat, wang2022fair}.

%% file: PAGES/conclusion.tex
\section{Conclusion}

In this work, we addressed the relationship between homophily and fairness, particularly regarding the OOD problem in the local homophily distribution of a graph. We provided a theoretical analysis of how the OOD problem affects predictions for nodes with different sensitive attributes. Then, using our three proposed fairness benchmarks and semi-synthetic graph generator, we conducted an empirical study to identify conditions that lead to unfairness in GNNs. Our findings showed the OOD problem is prevalent and most severe in graphs with high global homophily, making it challenging to integrate heterophilous users into homophilous settings. Although fair GNN models can partially address this issue, significant gaps persist. This work shifts the focus from addressing sensitive attributes to also considering the structural under-representation in GNN fairness. Future research will aim to enhance fairness for all users across the network, rather than just the majority.

%% file: PAGES/appendix.tex
\appendix
\section{Appendix}
\subsection{Proof for Theorem \ref{theorem:diff_in_sens_logit}}

\label{proof}

As outlined in the main text, we aim to study the OOD problem by "training" a GNN model under biased conditions, and then analyzing the resultant predictions for OOD nodes. 
The set of possible feature vectors for the different label and sensitive attributes are: 

\[
\mathbf{x}_i = 
\begin{cases}
   [-p_i, -q_i] \quad \text{if }  y = 0, s = 0\\
   [p_i,-q_i] \quad \text{if }  y = 1, s = 0\\
   [- p_i,   q_i] \quad \text{if }  y = 0, s = 1\\
   [p_i,  q_i] \quad \text{if }  y = 1, s = 1\\
\end{cases}
\]
where $p_i \overset{\mathrm{i.i.d}}\sim \mathcal{N}(\mu_{l}, \sigma_{l})$ and $q_i \overset{\mathrm{i.i.d}}\sim  \mathcal{N}(\mu_{s}, \sigma_{s})$. In the case that the $\mu_{l}$ or $\mu_{s}$ are close to 0, the features will follow similar distributions, and act as noise to the GNN. When $\mu_{l}$ or $\mu_{s}$ are larger, the vectors will more explicitly encode both the label and sensitive attribute. We note that we assume small values for $\sigma_{l}$ and $\sigma_{s}$ relative to $\mu_{l}$ and $\mu_{s}$ to mitigate the vectors becoming noise due to $\sigma_{l}$ and $\sigma_{s}$ overpowering $\mu_{l}$ and $\mu_{s}$. Since $\sigma_{l}$ and $\sigma_{s}$ are not the focus of this work, we let $\sigma_{s} = \sigma_{l} = \sigma$ for brevity. 

The biased conditions are induced by training the GNN model using $n$ data points, where the data points either have $y = s = 0$ or $y = s = 1$. Note that the GNN could also have been trained with $y = 0,  s = 1$ and $y = 1, s = 0$ through a simple notation swap, and this specific choice does not impact the results. We also assume that there are $k$ points where $ y = s = 0$, and $n - k$ points where $y = s = 1$. Using the GNN architecture $\mathbf{(A+I)XW}$, where $\mathbf{A+I} \in \{1, 0\}^{n \times n}$, $\mathbf{X} \in \mathbb{R}^{n \times 2}$, and $\mathbf{W} \in \mathbb{R}^{2 \times 2}$ the representations for each node after aggregation (without the linear transformation of $\mathbf{W}$) are:
\begin{equation}
\begin{aligned}
\mathbf{r}_{y=s=0} = \mathbf{x}_{y=s=0}  + hd\mathbf{x}_{y=s=0} + (1-h)d\mathbf{x}_{y=s=1}
\end{aligned}
\label{r00}
\end{equation}
\begin{equation}
\begin{aligned}
\mathbf{r}_{y=s=1} = \mathbf{x}_{y=s=1}  + hd\mathbf{x}_{y=s=1} + (1-h)d\mathbf{x}_{y=s=0}
\end{aligned}
\label{r11}
\end{equation}
where $d$ is the degree and $h$ is the homophily level. Thus, each representation is a weighted sum of the homophilous and heterophilous connections. 
Note that for connectivity we assume that the homophily level holds true for both class and sensitive attribute, i.e. homophilous connections will be homophilous with respect to both class and sensitive attribute, while the same property holds true for heterophilous connections.  We then establish the matrix $\mathbf{R}$ which holds the individual $r$ vectors in \eqref{r00} and \eqref{r11}.

We next find the weight matrix $\mathbf{W}$ that solves $\mathbf{RW = Y}$. However, there are two challenges: (1) As $\mathbf{R}$ is not a square matrix, $\mathbf{R}$ cannot be directly inverted to isolate $\mathbf{W}$, and (2) $\mathbf{R}$ possesses elements which are random variables. To handle the inversion process, we utilize a left pseudo-inverse through $\mathbf{W = (R^{T}R)}^{-1}\mathbf{R^{T}Y}$ to attain a solution to the system of equations. Then, we solve for the expected $\mathbf{W}$ by $\mathbf{\mathbb{E}[W] = \mathbb{E}[(R^{T}R)}^{-1}\mathbf{R^{T}Y}]$. As the random variables are uncorrelated, we can use the linearity of expectation to simplify the RHS, where

\begin{equation}
   \begin{split}
        \mathbb{E}[\mathbf{W}]
        &= \mathbb{E}[\mathbf{(R^{T}R)}^{-1}\mathbf{R^{T}}\mathbf{Y}] \\
        &= \mathbb{E}[\mathbf{(R^{T}R)}^{-1}]\mathbf{ \mathbb{E}[\mathbf{R^{T}}] \mathbf{Y}} \\
        &= \mathbb{E}\mathbf{[(R^{T}R)}^{-1}] \mathbb{E}[\mathbf{X^T\tilde{A}}] \mathbf{Y} \\
   \end{split}
\end{equation}

We use $\mathbf{\tilde{A}}$ to denote $\mathbf{A+I}$. Additionally, $\mathbf{\tilde{A} = \tilde{A}^{T}}$ since the graph is undirected. 
The second expectation is relatively straightforward as $\mathbf{\tilde{A}}$ can be pulled out, leaving an expectation over $\mathbf{X^T}$. Then, the element-wise application of expectation results in a new matrix, $\mathbf{\bar{X}^{T}}$ where each random variable is replaced with the mean of their respective distribution. For example, $x_i = [-\mu_{l}, - \mu_{s}]$ when y = 0, s = 0. For the first term, we compute the two different possible scenarios for $\mathbf{R}$: 

\begin{equation}
\mathbf{r}_i = 
\begin{cases}
   -(dh + 1)[p_i, q_i] + d(1-h)[p_i, q_i] &\quad \text{if }  y, s = 0\\
   (dh + 1)[p_i,  q_i] - d(1-h)[p_i, q_i] &\quad \text{if }  y, s = 1\\
\end{cases}
\end{equation}
This can be represented in matrix form as $\mathbf{R} = (2dh - d + 1) \mathbf{D G}$, where $\mathbf{D} = \text{diag}([-1, ..., -1, ..., 1, ..., 1])$ is an $n \times n$ diagonal matrix and $\mathbf{G} \in \mathbb{R}^{n \times 2}$ is a Gaussian random matrix where each element $\mathbf{G}_{i, 0} \sim \mathcal{N}(\mu_{l},\sigma)$ and $\mathbf{G}_{i, 1} \sim \mathcal{N}(\mu_{s},\sigma)$. Then, 

\begin{equation}
    \begin{split}
        \mathbf{({R^TR})}^{-1} &= (2dh - d + 1)^{-2}(\mathbf{(DG)^T(DG))}^{-1} \\
                                      &= (2dh - d + 1)^{-2}(\mathbf{(G^{T}DDG)}^{-1}
    \end{split}
\end{equation}
Since $\mathbf{D}$ is a diagonal matrix, $\mathbf{D^T = D}$. Moreover, $\mathbf{DD = I}$, thus $\mathbf{({R^TR})^{-1}} = (2dh - d + 1)^{-2}(\mathbf{G^TG})^{-1}$. Then, letting $\mathbf{\bar{R}} = (\mathbf{\bar{X}^{T}\tilde{A})^{T}} $, i.e. the aggregation of expected elements within $\mathbf{X}$, 

\begin{equation}
   \begin{split}
        \mathbb{E}[\mathbf{W}]
        &= (2dh - d + 1)^{-2}\mathbb{E}\mathbf{[(\mathbf{G^TG})^{-1}}] \mathbf{\bar{R}^{T}} \mathbf{Y} \\
   \end{split}
\end{equation}

Explicitly calculating $\mathbb{E}\mathbf{[(\mathbf{G^TG})^{-1}}]$ can be difficult due to the dependence induced by the off-diagonal terms of $
\mathbf{G^TG}$. Instead, we will approximate this relationship by using a first-order Taylor Expansion of $\mathbf{(\mathbf{G^TG})^{-1}}$. Specifically, we know if we have a function $f(\mathbf{M}) = (\mathbf{M^{T}M})^{-1}$, then 
\[f(\mathbf{M}) \approx f(\mathbf{M_{0}}) + \dfrac{\text{d}f}{\text{d}\mathbf{M}}(\mathbf{M_{0})}\cdot (\mathbf{M - M_{0})}, \]
where $\mathbf{M_{0}}$ is the point $f(\mathbf{M})$ is being approximated at. Approximating at $\mathbf{\bar{G}}$, the expected matrix of $\mathbf{G}$,

\begin{equation}
\begin{split}
    f(\mathbf{G}) &\approx (\mathbf{\bar{G}^T\bar{G})}^{-1} + \dfrac{\text{d}f}{\text{d}\mathbf{G}}(\mathbf{\bar{G})}\cdot(\mathbf{G - \bar{G}}) \\
    \mathbb{E}[f(\mathbf{G})] &\approx  \mathbb{E}[(\mathbf{\bar{G}^T\bar{G})}^{-1} + \dfrac{\text{d}f}{\text{d}\mathbf{G}}(\mathbf{\bar{G})}\cdot(\mathbf{G - \bar{G}})]    \\
    &\approx  \mathbb{E}[(\mathbf{\bar{G}^T\bar{G})}^{-1}] + \mathbb{E}[\dfrac{\text{d}f}{\text{d}\mathbf{G}}(\mathbf{\bar{G})}\cdot(\mathbf{G - \bar{G}})]
\end{split}
\end{equation}

As the first term is constant, the expectation is simply $(\mathbf{\bar{G}^T\bar{G})}^{-1}$. Simplifying the second term, using $\dfrac{\text{d}f}{\text{d}\mathbf{M}}(\mathbf{M}^{-1}) = -\mathbf{M}^{-1} (\dfrac{\text{d}}{\text{d}\mathbf{M}} (\mathbf{M^T}\mathbf{M})) \mathbf{M}^{-1} = -\mathbf{M}^{-1} (\text{d}\mathbf{M^T}\mathbf{M} + \mathbf{M^T}\text{d}\mathbf{M})) \mathbf{M}^{-1} $, 

\begin{equation}
\begin{split}
    \mathbb{E}[f(\mathbf{G})] &\approx (\mathbf{\bar{G}^T\bar{G})}^{-1} \\
    &+ \mathbb{E}[-\mathbf{\bar{G}^T\bar{G}}^{-1} (\text{d}\mathbf{G^T}\mathbf{\bar{G}} + \mathbf{\bar{G}^T}\text{d}\mathbf{G}) \mathbf{(\bar{G}^T\bar{G}})^{-1} ] 
\end{split}
\end{equation}

Using the fact that we can decouple $\mathbf{G}$ as a matrix of means, and matrix of elements sampled from a zero-mean normal distribution, i.e. $\mathbf{G = \bar{G} + }\text{d}\mathbf{G} $, we can rewrite the differential as $\text{d}\mathbf{G} = \mathbf{G - \bar{G}}$. Thus, $\text{d}\mathbf{G}$ is a matrix of zero-mean Gaussian random variables, leading to an expectation of zero in the second term as seen below.

\begin{equation}
\begin{split}
    \mathbb{E}[-\mathbf{\bar{G}^T\bar{G}}^{-1} (\text{d}\mathbf{G^T}\mathbf{\bar{G}} + \mathbf{\bar{G}^T}\text{d}\mathbf{G}) \mathbf{(\bar{G}^T\bar{G}})^{-1} ] = \\
    -\mathbf{\bar{G}^T\bar{G}}^{-1} \mathbb{E}[(\text{d}\mathbf{G^T}\mathbf{\bar{G}} + \mathbf{\bar{G}^T}\text{d}\mathbf{G})] \mathbf{(\bar{G}^T\bar{G}})^{-1} = 0
\end{split}
\end{equation}
Thus, $\mathbb{E}[f(\mathbf{G})] \approx (\mathbf{\bar{G}^T\bar{G})}^{-1}$, leading to,

\begin{equation}
   \begin{split}
        \mathbb{E}[\mathbf{W}]
        &\approx (2dh - d + 1)^{-2} (\mathbf{\bar{G}^T\bar{G})}^{-1} \mathbf{\bar{R}^{T}} \mathbf{Y} \\
   \end{split}
\end{equation}
Computing \(\mathbf{\bar{G}^T \bar{G}}\):

\[
\mathbf{\bar{G}^T \bar{G}} = 
\begin{pmatrix}
\mu_l & \cdots & \mu_l \\
\mu_s & \cdots & \mu_s 
\end{pmatrix} 
\begin{pmatrix}
\mu_l & \mu_s \\
\vdots & \vdots \\
\mu_l & \mu_s 
\end{pmatrix} 
= n
\begin{pmatrix}
\mu_l^2 & \mu_l\mu_s \\
\mu_l\mu_s & \mu_s^2
\end{pmatrix}
\]

To ensure that \(\mathbf{\bar{G}^T \bar{G}}\) can be inverted, we add a Tikhonov regularization term. Thus, 

\[
\mathbf{\bar{G}^T \bar{G}} \approx  \mathbf{\bar{G}^T \bar{G} + \lambda \mathbf{I}} 
= n
\begin{pmatrix}
\mu_l^2 + \frac{\lambda}{n} & \mu_l\mu_s \\
\mu_l\mu_s & \mu_s^2 + \frac{\lambda}{n} 
\end{pmatrix}
\]
When $\lambda > 0$, $\det(\mathbf{\bar{G}^T \bar{G} + \lambda \mathbf{I}}) \ne 0$, and then

\begin{equation}
   \begin{split}
        \mathbb{E}[\mathbf{W}]
        \approx \frac{(2dh - d + 1)^{-2}}{(n\mu_l^2 + \lambda)(n\mu_s^2 + \lambda) - (n\mu_l\mu_s)^2}  \\
        \begin{pmatrix}
        n\mu_s^2 + \lambda & -n\mu_l\mu_s \\
        -n\mu_l\mu_s & n\mu_l^2 + \lambda
        \end{pmatrix}
\mathbf{\bar{R}^{T}} \mathbf{Y} \\
   \end{split}
\end{equation}

Additionally, we can simplify $\mathbf{\bar{R}^{T}}$. Assuming that the first $k$ rows are data points where $y = s = 0$ and the bottom $n - k$ rows are data points where $y = s = 1$, 

\begin{equation}
   \begin{split}
        \mathbf{\bar{R}^{T}} \mathbf{Y} & = (2dh - d + 1)\mathbf{\bar{G}^T D} \mathbf{Y} \\
        &= (2dh - d + 1)\begin{pmatrix}
        -k\mu_l  & (n-k)\mu_l \\
        -k\mu_s  & (n-k)\mu_s 
        \end{pmatrix}
   \end{split}
\end{equation}
Thus,
\begin{equation}
   \begin{split}
        \mathbb{E}[\mathbf{W}]
        \approx \frac{(2dh - d + 1)^{-1}}{(n\mu_l^2 + \lambda)(n\mu_s^2 + \lambda) - (n\mu_l\mu_s)^2}  \\
        \begin{pmatrix}
        n\mu_s^2 + \lambda & -n\mu_l\mu_s \\
        -n\mu_l\mu_s & n\mu_l^2 + \lambda
        \end{pmatrix}
        \begin{pmatrix}
        -k\mu_l  & (n-k)\mu_l \\
        -k\mu_s  & (n-k)\mu_s 
        \end{pmatrix}
   \end{split}
\end{equation}

We now consider how test data points with shifted local homophily level are predicted under the learned weight matrix. Without loss of generality, we assume the test data points have label $y = 0$, and study how the change in local homophily level can induce disparity between sensitive attributes. We define two new aggregated feature vectors following the assumptions above for $y = 0$: 
\begin{equation}
\begin{aligned}
\mathbf{r}_{y=s=0} = \mathbf{x}_{y=s=0}  + (h+\alpha)d\mathbf{x}_{y=s=0} + \\(1-(h+\alpha))d\mathbf{x}_{y=s=1}
\end{aligned}
\label{r00a}
\end{equation}

\begin{equation}
\begin{aligned}
\mathbf{r}_{y=0, s=1} = \mathbf{x}_{y=0, s=1}  + (h+\alpha)d\mathbf{x}_{y=0, s=1} + \\ (1-(h+\alpha))d\mathbf{x}_{y=1, s=0}
\end{aligned}
\label{r01a}
\end{equation}
where $\alpha$ denotes a shift in the local homophily level for the test node. Then, we analyze two scenarios: $\alpha = 0$ and $\alpha \ne 0$. This leads to the to the set of vectors,

\begin{equation}
\mathbf{r}_{t} = 
\begin{cases}
   \begin{aligned}
      -(d(h + \alpha) + 1)[p_i, q_i] + &\\
      d(1-(h + \alpha))[p_i, q_i] &
   \end{aligned} &\quad   s = 0 \\
   \begin{aligned}
      (d(h + \alpha) + 1)[-p_i,  q_i] + &\\
      d(1-(h + \alpha))[p_i, -q_i] &
   \end{aligned} &\quad s = 1\\
\end{cases}
\end{equation}

We get the expected prediction for each test data point $t$ as $\mathbb{E}[\mathbf{p}_{t}] = \mathbb{E}[\mathbf{r}_{t}]\mathbb{E}[\mathbf{W}]$ given they are independent. 
In the following results, we use $u$ to represent the test point where $s=0$ case, and $v$ to represent the test point where $s=1$ to simplify notation. To assess how the expected predictions vary, we compute the difference between the expected predictions of $u$ and $v$ as a function of $\alpha$, where

\begin{equation}
    \begin{split}
            \mathbb{E}[\mathbf{p}_{u, \alpha}] - \mathbb{E}[\mathbf{p}_{v, \alpha}] = \\ 
            \frac{\mu_{s}^{2}k(1+d(2h + 2\alpha - 1))}{(1 + 2dh-d)(\lambda + (\mu_{l}^2 + \mu_{s}^2)n}
    \end{split}
    \label{eqn_diff}
\end{equation}
Thus, we have the expected difference in predictions as a function of $\alpha$. \hfill $\square$

\subsection{Deeper Discussion on Previous Fairness Benchmarks}
\label{prev_datasets}
Below we further expand on the points raised in the main text regarding fairness benchmarks.

\vspace{0.1cm}
\noindent \textbf{Inadequate Range of Local Homophily Patterns: } 
Earlier works proposed the German, Recidivism, and Credit similarity networks as fairness benchmarks \cite{agarwal2021towards}. These datasets are constructed by computing feature similarities within non-graph datasets, resulting in highly homophilous structures given the correlation between features, structure, and labels. While these datasets have become common fairness benchmarks \cite{chen2024fairness, dong2023elegant}, their generally non-varied structure provides minimal insight into how fairness can vary across different homophily levels. Consequently, fairness results on these datasets overlook OOD-induced unfairness and can only assess how homophilous message passing amplifies sensitive attribute-label bias. 

\vspace{0.1cm}
\noindent \textbf{Reproducibility Challenges: } 
Disparities affecting minority groups may be sensitive to data splitting and processing choices, necessitating complete access to datasets for retrospective studies on fair GNNs \cite{ma2022subgroupfair}. Unfortunately, some datasets lack detailed information on subgraph sampling, as well as sensitive attribute choices \cite{dong2023fairness}. For instance, the Pokec fairness benchmark provides labels and sensitive attributes for only about 1\% of the nodes, with no mechanism to recover missing information \cite{dai2021fairgnn}. Similarly, the Facebook social networks for college campuses suffer from ambiguity, with multiple dataset versions existing \cite{dong2023fairness, chen2024fairness}. The Facebook datasets also face challenges in fairness studies given their canonical task is to predict the gender of a user, with no other clear sensitive attribute. Without the full graph, as well as a clearly defined sensitive attribute, deciphering biases is extremely difficult, necessitating a standardization of the data generation process.  

\vspace{0.1cm}
\noindent \textbf{Unclear Fairness Implications: } To bolster the number of fairness benchmarks, datasets without sensitive attributes or fairness contexts have been adopted from the GNN literature~\cite{chen2024fairness}. For instance, WebKB datasets like Squirrel and Chameleon have been used for fairness studies. However, the sensitive attributes are chosen as structural attributes of the network, rather than from the feature set. From the perspective of fairness, sensitive attributes are defined as characteristics which have driven disparate treatment of individuals \cite{zhao2022towards, dong2023fairness}. Moreover, as  fairness for GNNs aims to study the interplay between individual attributes and graph structure, defining the sensitive attribute relative to structure is a fundamentally different task and obscures the difference between sensitive attribute fairness and structure-performance relationships in GNNs.

\subsection{Training Details and Hyperparameters}

\label{app:training_details}

All models are implemented within PyTorch Geometric, with the fair GNNs from PyGDebias \cite{dong2023fairness}. 
We use the Adam optimizer (torch.optim.adam), searching over learning rates ${0.001, 0.0001, 0.00001}$. The weight decay is 0.0. Each model is trained on a single NVIDIA Volta V100 GPU.

\noindent \textbf{GCN} (torch\_geometric.nn.conv.gcn\_conv): a) Hidden Dim:~32, 64. b) Depth: 2, 3. c) Dropout: 0.3, 0.5.

\noindent \textbf{SAGE} (torch\_geometric.nn.conv.sage\_conv): a) Hidden Dim:~32, 64. b) Depth: 2, 3. c) Dropout: 0.3, 0.5.

\noindent \textbf{LINKX} (torch\_geometric/nn/models/LINKX): a) Hidden Dim:~32, 64. b) Depth: 2, 3. c) Dropout: 0.3, 0.5.

\noindent \textbf{FairGNN} (\href{https://github.com/yushundong/PyGDebias/blob/main/pygdebias/debiasing/FairGNN.py}{PyGDebias/FairGNN}): a) Hidden Dim: 32, 64. b) Depth: 2, 3. c) Dropout: 0.3, 0.5. The tradeoff parameters $\alpha$ and $\beta$ are set to 0.01 and 5 as recommended in \cite{dai2021fairgnn}.

\noindent \textbf{Nifty} (\href{https://github.com/yushundong/PyGDebias/blob/main/pygdebias/debiasing/NIFTY.py}{PyGDebias/Nifty}): a) Hidden Dim: 32, 64. b) Depth: 2, 3. c) Dropout: 0.3, 0.5. The tradeoff parameter $\lambda$ is set to 0.6, as recommended in \cite{agarwal2021towards}.

\subsection{Semi-Synthetic Psuedocode}

\label{semisynth_psuedo}
In algorithm \ref{alg:alter_local_homophily} we provide pseudo-code for the generation of the semi-synthetic graphs. This is shown for only the re-wiring portion of the generation process, but the addition process follows similar logic without the removal step. The full code can be found at \href{https://anonymous.4open.science/r/HeteroFairICDM-F7DE/synthetic_datasets/generate_synth_data_rewire.py}{Anonymous Repo}. As input to the algorithm, we assume the transportation matrix and edge editing bounds are pre-computed. For instance, the solution for the transportation matrix is generated using the POT library \cite{flamary2021pot}.

\subsubsection{Notation}

To clarify notation, the method assumes as input a graph G, the proportion matrix learned during the optimal transport optimization $\mathbf{T}$, the pre-computed initial local homophily levels $\mathbf{h}$, and the number of discrete bins $b$. As there are more bins, the semi-synthetic data will be able to better approximate the goal distribution. Additionally, we use the sgn equation to denote whether the change is positive or negative. This information allows us to determine whether a node needs to become more homophilous or heterophilous. 

\begin{algorithm}[H]
\caption{Re-wiring Edges}
\label{alg:alter_local_homophily}
\begin{algorithmic}[1]
\State \textbf{Input:} Graph $G = (V, E)$, proportions matrix $\mathbf{T}$, local homophily vector $\mathbf{h}$, number of bins $b$
\State \textbf{Output:} Modified graph $G' = (V, E')$

\State Sample nodes $S$ from $V$ based on proportions in $\mathbf{T}$ \Comment{Source node sampling}

\For{each node $v_i \in S$}
    \State Get number of edges to modify $m_i$
    \State $diff = \text{sgn}(h_{i, g} - h_{i, c})$
    \If{$diff > 0$} \Comment{Add homophily}
        \State Sample up to $m_i$ heterophilous edges from $v_i$ to $v_{j}$ if $\text{sgn}(h_{j, g} - h_{j, c}) = diff$
    \Else \Comment{Add Heterophily}
        \State  Sample up to $m_i$ homophilous edges from $v_i$ to $v_j$ if $\text{sgn}(h_{j, g} - h_{j, c}) \ne diff$
    \EndIf
    \For{each edge $(v_i, v_j)$}
        \State Select $v_k \notin N(v_i)$ s.t. $\text{sgn}(h_{k, g} - h_{k, c}) = diff$ 
        \State $G \leftarrow (G \setminus (v_i, v_j)) \cup (v_i, v_k)$
    \EndFor
\EndFor

\State \textbf{Return} modified graph $G$
\end{algorithmic}
\end{algorithm}

\subsection{EMD for Semi-Synthetic Data}

In table \ref{table:synth_eval}, we provide the Earth Mover's Distance (EMD) between the goal and original, as well as goal and generated local homophily distributions. For instance, the unmodified local homophily distribution of Tolokers had an EMD of 0.41 with the goal distribution specified by $\alpha=3$ and $\beta=10$. Then, after re-wiring, the distance became 0.06.

\begin{table}[!htbp]
\centering
\caption {EMD between Goal and Original / Gen \\ Local Homophily Distribution.}
\begin{tabular}{@{}lll@{}}
\toprule
    & \multicolumn{2}{l}{Beta Shape Parameters} \\ \cmidrule(l){2-3}
    & 
    $\alpha$=3, $\beta$=10         & $\alpha$=10, $\beta$=3        \\ \midrule
\textbf{Tolokers}   &  0.41 / 0.06     &  0.14 /  0.02     \\
\textbf{Pokec} &  0.16 / 0.10     &  0.38 / 0.15    \\
\textbf{FB-Penn94} &  0.18 / 0.04    &  0.37 /  0.03    \\ \bottomrule
\end{tabular}
\label{table:synth_eval}
\end{table}